\documentclass[runningheads]{llncs}
\usepackage{graphicx}

\usepackage{tikz}
\usepackage{comment}
\usepackage{amsmath,amssymb} 
\usepackage{multirow}
\usepackage{floatrow}
\usepackage{color}
\usepackage{xcolor}

\usepackage{float}
\floatstyle{plaintop}
\restylefloat{table}

\usepackage[T1]{fontenc}
\usepackage[utf8]{inputenc}
\usepackage[font=small,labelfont=bf]{caption}
\usepackage{subcaption}

\definecolor{citecolor}{HTML}{0071bc}
\usepackage[pagebackref=true,breaklinks=true,letterpaper=true,colorlinks,citecolor=citecolor,bookmarks=false]{hyperref}

\newfloatcommand{capbtabbox}{table}[][\FBwidth]

\usepackage{blindtext}
\usepackage{amssymb,algorithm}
\usepackage{algpseudocode}
\newcommand{\etal}{\textit{et al.}}
\usepackage[font=small,labelfont=bf]{caption}

\usepackage[accsupp]{axessibility}  
\usepackage{cite}

\newlength\savewidth\newcommand\shline{\noalign{\global\savewidth\arrayrulewidth
  \global\arrayrulewidth 1pt}\hline\noalign{\global\arrayrulewidth\savewidth}}

\DeclareMathOperator*{\argmax}{arg\,max}

\newcommand{\ym}{y_{\text{mask}}}
\newcommand{\ykp}{y_{\text{kp}}}
\newcommand{\mm}{\mu_{\text{mask}}}
\newcommand{\mkp}{\mu_{\text{kp}}}

\newcommand{\tkp}{\theta_{\text{kp}}}
\newcommand{\tq}{\theta_{\text{q}}}

\newcommand{\lm}{\ell_{\text{mask}}}
\newcommand{\lkp}{\ell_{\text{kp}}}
\begin{document}

\pagestyle{headings}
\mainmatter
\def\ECCVSubNumber{1408}  

\title{Improving Few-Shot Part Segmentation using Coarse Supervision} 


\titlerunning{Improving Few-Shot Part Segmentation using Coarse Supervision}
%
\author{Oindrila Saha \quad \quad Zezhou Cheng \quad \quad  Subhransu Maji\\
\institute{University of Massachusetts, Amherst}
{\tt\small \{osaha, zezhoucheng, smaji\}@cs.umass.edu}
}

\authorrunning{O. Saha et al.}

\maketitle
\begin{abstract}
A significant bottleneck in training deep networks for part segmentation is the cost of obtaining detailed annotations.
We propose a framework to exploit coarse labels such as figure-ground masks and keypoint locations that are readily available for some categories to improve part segmentation models.
A key challenge is that these annotations were collected for different tasks and with different labeling styles and cannot be readily mapped to the part labels.
To this end, we propose to jointly learn the dependencies between labeling styles and the part segmentation model, allowing us to utilize supervision from diverse labels.
To evaluate our approach we develop a benchmark on the Caltech-UCSD birds and OID Aircraft dataset.
Our approach outperforms baselines based on multi-task learning, semi-supervised learning, and competitive methods relying on loss functions manually designed to exploit coarse supervision. 
\end{abstract}


\section{Introduction}
Accurate models for labeling parts of an object can aid fine-grained recognition tasks such as estimating the shape and size of animals, and support applications in graphics such as image editing and animation.
But a significant bottleneck is the cost of collecting annotations for supervising part labeling models.
In many situations however, one can find datasets with alternate labels such as object bounding boxes, figure-ground masks, or keypoints, which may serve as a source of supervision.
However the variations in their level of detail and structure, e.g., bounding boxes and masks are coarser than part labels while keypoints are sparse, implies that they cannot be readily ``translated'' to part labels to directly supervise learning. 

\begin{figure*}[t]
    \centering
    \includegraphics[width=0.9\linewidth]{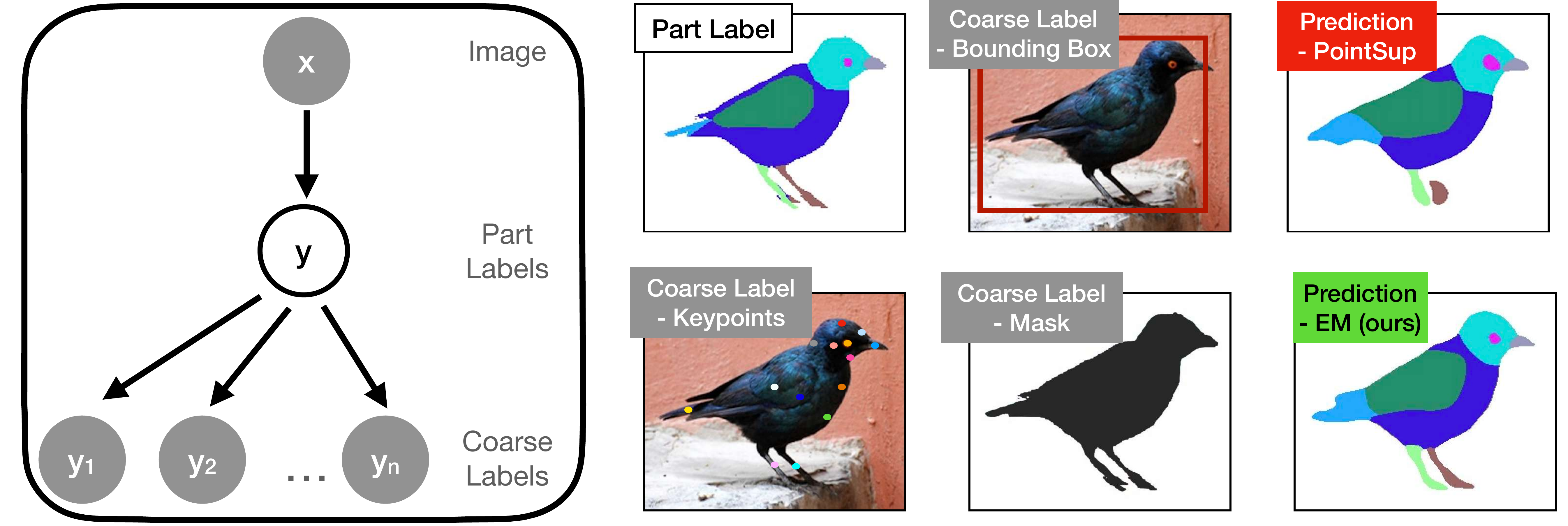}
    \caption{\small{\textbf{Overview of our approach.} The graphical model over an image $x$, parts labels $y$ and coarse labels $y_1, \ldots, y_n$, is shown on the left. Coarse labels such as bounding boxes, figure-ground masks, or keypoint locations are easier to annotate than per-pixel part labels, and our learning framework can utilize datasets with coarse labels to train part segmentation models, outperforming previous work - PointSup (Fig.7).}}
    \label{fig:splash}
\end{figure*}

We propose a framework to learn part segmentation models using existing datasets with coarse labels such as figure-ground masks and keypoints. 
The approach illustrated in Fig.~\ref{fig:splash} treats part labels as latent variables and jointly learns the part segmentation model and the \emph{unknown} dependencies between the labeling styles in a Bayesian setting (\S~\ref{sec:method}). 
The dependencies are represented using deep networks to model complex relationships between labeling styles, allowing supervision from a variety of coarse labels. 
One technical challenge is that inference requires sampling over high-dimensional latent distributions which is typically intractable. 
We address this by making certain conditional independence assumptions and develop an amortized inference procedure for learning. Our method allows the use off-the-shelf  image segmentation networks and standard back-propagation machinery for training.

To evaluate our approach we design a benchmark for labeling parts on the Caltech-UCSD birds (CUB)~\cite{WahCUB_200_2011} and OID Aircraft~\cite{mahendran14understanding} dataset (\S~\ref{sec:benchmark}). 
We utilize the keypoint and masks from the CUB dataset and the part-segmentation labels of the birds of PASCAL VOC to segment birds into 10 parts. 
Our approach achieves a performance of 49.25\% mIoU compared to baseline of fine-tuning an ImageNet pre-trained network on all the available part labels (45.37\% mIoU), as well as multi-tasking (41.27\% mIoU) and semi-supervised learning baseline (46.01\% mIoU). 
It also outperforms PointSup~\cite{cheng2021pointly} (46.76\% mIoU), an approach for training using point-supervision -- for this approach we manually assign keypoints to parts and combine it with the figure-ground mask to provide a set of part segmentation labels. 
On the OID Aircraft dataset we observed a similar trend, where our approach (58.68\% mIoU) outperforms the fine-tuning (55.3\% mIoU) and multi-tasking (55.61\% mIoU) baselines.
These experiments are consistent across different initializations of the network (e.g., ImageNet pre-trained vs. random), as well as the types and combinations of coarse labels (\S~\ref{sec:algorithms} \& \S~\ref{sec:results}).

Our approach is also relatively efficient --- fine-tuning on all the labeled parts of the CUB dataset requires 1 hour, PointSup~\cite{chen2014detect} requires 2.5 hours, while our approach requires 7.5 hours on a single NVIDIA RTX-8000 GPU.
Importantly, our approach requires little additional labeling ($\approx$300 or $<$3\% instances for CUB dataset), and benefits from \emph{existing} part labels on PASCAL and coarse labels on CUB.
These experiments suggest that diverse coarse labels across datasets can be used to effectively guide part labeling tasks within our framework.

To summarize our contributions include: 1) a framework to learning part segmentation models using diverse coarse supervision from existing datasets; 2) an amortized inference procedure that is efficient, roughly 3$\times$ slower than the leading alternate methods for coarse supervision (e.g., PointSup~\cite{cheng2021pointly}), and is more accurate; 3) two benchmarks for evaluating part-segmentation from a few-labeled examples on the CUB and OID Aircraft dataset; and 4) a systematic evaluation of various design choices including the role of initialization for transfer learning and the relative benefits of various forms of coarse labels. The source code and data associated with the paper are publicly available at \href{https://people.cs.umass.edu/~osaha/coarsesup}{https://people.cs.umass.edu/$\sim$osaha/coarsesup}


\section{Related work}
\paragraph{\textbf{Weakly supervised image segmentation}} Previous work use supervision from classification labels, bounding boxes or at sparse locations in the image such as points or lines. Zhou \etal~\cite{zhou2018weakly} use class response at every image location for a given class and train by mapping the response peaks to more informative parts of an object instance. Other approaches generate pseudo ground truth labels using previous image classification models~\cite{ahn2019weakly, zhu2019learning}. Khoreva \etal~\cite{khoreva2017simple} use bounding boxes as weak supervision. They generate pseudo ground truth using classical approaches such as GrabCut~\cite{rother2004grabcut} inside given bounding box and use that to train the segmentation model. Hsu \etal~\cite{hsu2019weakly} use a bounding box tightness prior and train a Mask-RCNN~\cite{he2017mask} using horizontal and vertical patches from the tight bounding box as positive signals and those outside as negative signals. BoxInst~\cite{tian2021boxinst} uses a projection loss that forces horizontal and vertical lines inside bounding boxes to predict at least one foreground pixel and an affinity loss that forces pixels with similar colors to have the same label. Laradji \etal~\cite{laradji2020proposal} introduce a proposal-based instance segmentation method that uses a single point per instance as supervision. Cheng \etal~\cite{cheng2021pointly} uses multiple points randomly sampled per instance as well as bounding boxes as supervision to train a Mask-RCNN model. ScribbleSup~\cite{lin2016scribblesup} uses a graphical model that jointly propagates information from scribbles to unmarked pixels to learn network parameters. Another stream of work~\cite{zou2020pseudoseg, chen2021semi} train two models simultaneously with cross supervision from one model to train the other. Naha \etal~\cite{naha2021part} use keypoint guidance to predict part segmentation labels for unseen classes but require keypoint inputs during evaluation time. All these methods design algorithms specific to one kind of supervision and the annotation style has a clear mapping to the desired part labels. \emph{In contrast, our method handles a variety of label styles allowing opportunistic use of existing datasets to learn part segmentation labels.}


\paragraph{\textbf{Unsupervised learning}}
A number of previous work use self-supervision for learning segmentations. 
SCOPS~\cite{hung2019scops} uses geometric concentration (areas of the same object part are spatially concentrated), equivariance (enforcing part segmentation to be aligned with geometric transformations) and semantic consistency (over different instances).
Wang \etal~\cite{wang2020self} also use equivariance constraints to refine class activation maps which in turn form the final segmentation maps. 
Another method~\cite{o2020unsupervised} uses pixel-level contrastive learning to learn feature representations for downstream tasks such as segmentation. 
Yang \etal~\cite{yang2021unsupervised} use a layered GAN to produce background and foreground layers for an image where the discriminator predicts on the overlayed image. 
PiCIE~\cite{cho2021picie} enforces invariance to photometric transformations and equivariance to geometric transformations for different views of the same image.
A number of recent techniques based on generative~\cite{zhang2021datasetgan, tritrong2021repurposing} and contrastive learning~\cite{liu2021few} approaches have also been proposed (See~\cite{saha2022ganorcon} for a systematic evaluation).
These methods can be used to initialize networks to boost performance in few-shot learning and are complementary to our approach. \emph{For example, we compare the benefits of self-supervised learning over randomly initialized networks and ImageNet pre-trained networks.}

\paragraph{\textbf{Multi-task learning}} benefits from diverse source of supervision by sharing parts of the model across tasks.
For image segmentation, a prior work~\cite{Dai_2016_CVPR} proposes multi-task cascaded networks where three networks predict instances, masks and categorize objects respectively.
Heuer \etal~\cite{Heuer_2021_ICCV} combine the tasks of object detection, semantic segmentation and human pose estimation but fails to perform better than the single task network for segmentation --- a trend we also observe in our experiments. 
Standley \etal~\cite{pmlr-v119-standley20a} show that combining some tasks in multi-task setting can degrade performance while for other cases performance can get boosted. To design a multi-task network able to handle different tasks, some methods~\cite{pmlr-v119-guo20e, fifty2021efficiently} group tasks that would perform  well together. Other works such as~\cite{tripathi2017pose2instance, Kocabas_2018_ECCV} use keypoints and bounding box information to predict instance segmentation but use a multistage framework. Mask-RCNN~\cite{he2017mask} adds a mask segmentation head to Faster-RCNN~\cite{ren2015faster} to predict bounding box and instance segmentation. 
\emph{Unlike generic multi-tasking approaches, our approach exploits the hierarchical label structure to guide learning and consistently outperforms them.}


\section{A Joint Model of Labeling Styles}\label{sec:method}

For an image $x$ denote $y \in S$ the part segmentation label, i.e., pixel-wise label for each part, and $y_1 \in S_1$, $y_2 \in S_2$, \ldots, $y_n \in S_n$ denote coarse labels corresponding to various labeling styles. 
For example, $y_1$ might denote the coordinates of a set of keypoints and $y_2$ might denote the figure-ground mask.
We call a labeling $S_a$ coarser than $S_b$ if $S_a$ can be derived from $S_b$ \emph{independent} of the image $x$. For example, the figure-ground mask can be derived from the part label of an object, or the bounding-box can be derived from the figure-ground mask. 
\emph{Our goal is to learn a part segmentation model $p(y|x)$ given a small set of images with part labels $y \in S$, and a large set of images with coarse labels $y_k \in S_{k}$.}


This assumption that the coarse labels can be derived from the part labels leads to the following joint probability distribution over the image and the labels:
\[
    p(y, y_1, \ldots, y_n|x) = p(y|x) \prod_{i=1}^n p(y_1|y), 
\]
and is illustrated by the graphical model in Fig.~\ref{fig:splash}. 
The assumption might appear to be strong, but we find that it holds for the styles of labels we consider. 
For example, a convolutional network can accurately predict the location of keypoints given the part segmentation labels with $>$ 92 PCK which is as good as the accuracy of keypoints given image.
However, the form of $p(y_k|y)$ is complex in this case as it involves reasoning about the extent and location of various parts. The distribution might also be unknown, especially when combining existing datasets which may have been collected with a different set of labels and annotation guidelines. For example, there might not be a direct correspondence between the names of parts used for keypoint annotations and those for segmentation task. 
In contrast, the form is simple and deterministic for figure-ground masks or bounding boxes given part labels. We incorporate this factorization in a Bayesian setup to learn both the part segmentation model and the dependencies between the labeling styles described next.


\subsection{Variational EM for Learning}
Assume that an image $x$ contains coarse labels $y_1, y_2, \ldots, y_n$. We will estimate parameters $\theta$ to maximize the log-likelihood of the data: 
\begin{equation}
\max_\theta{\cal L}(\theta) = \log p(y_1, y_2, \ldots, y_n | x, \theta). 
\end{equation}
Given a distribution $q(y)$ over the latent variables\footnote{$q(y) \geq 0$ and $p(y, y_1, y_2, \ldots, y_n) > 0 \Rightarrow q(y) > 0$} the ${\cal L}(\theta)$ can be bounded as:
\begin{equation} \label{eq:joint}
\begin{aligned}
{\cal L}(\theta) &= \log \sum_{y} p(y, y_1, y_2, \ldots, y_n | x, \theta) \\
&=  \log \sum_{y} q(y) \frac{p(y, y_1, y_2, \ldots, y_n | x, \theta)}{q(y)} \\
&\geq  \sum_{y} q(y) \log  \frac{p(y, y_1, y_2, \ldots, y_n | x, \theta)}{q(y)} \\
&=   \sum_{y}  q(y)\log p(y, y_1, y_2, \ldots, y_n | x, \theta) + H(q) \\
&=   \sum_{y}  q(y) \left[ \log p(y|x)\prod_{i=1}^n p(y_i|y, \theta) \right] + H(q) := {\cal F}(q, \theta).\\
\end{aligned}
\end{equation}
Where $H(q) = -\sum_y q(y) \log q(y)$ is the entropy of the distribution $q$. The EM algorithm alternates between:
\begin{itemize}
\item \textbf{E step:} maximize ${\cal F}(q,\theta)$ wrt distribution over $y$ given the parameters:
    \[
        q^{(k)}(y) = \argmax_{q(y)} {\cal F}(q(y), \theta^{(k-1)}). 
    \]
\item \textbf{M step:} maximize ${\cal F}(q,\theta)$ wrt parameters given the distribution $q(y)$:
    \[
        \theta^{(k)} = \argmax_{\theta} {\cal F}(q^{(k)}(y), \theta) = \argmax_{\theta} \sum_y q^{(k)}(y) \log p(y, y_1, y_2, \ldots y_n| x, \theta) 
    \]
\end{itemize}
Note that in the above we have derived the EM algorithm for a single example $x$, but the overall approach requires estimating the distribution over latent variables $q(y)$ for each training example and parameters across all the training examples. 
However, optimizing $q(y)$ for each training sample $x$ is typically intractable for high-dimensional distributions like ours. In ``Hard EM" the distribution $q(y)$ is replaced by the mode of the posterior distribution but estimating this can also be challenging when the probabilities are expressed using deep networks.
In the next section we present an amortized inference procedure where we estimate $q(y)$ using a separate network conditioned on all the observed variables.

\subsection{Coarse Supervision from Keypoints and Figure-Ground Mask}
As a concrete example consider that two types of coarse labeling styles are available --  $y_\text{mask} \in S_\text{mask}$ denoting the figure-ground mask of the same size as the image, and $y_\text{kp} \in S_\text{kp}$ denoting the locations of a set of keypoints in an image. To make inference tractable we adopt a Laplace approximation and model the conditional distributions as a random variable centered around a mean as follows:
\begin{itemize}
    \item $p(y|x) \propto \exp(-\alpha|y - \mu(x)|) $ where $\mu(x)$ is the mean distribution of the part labels for the image estimated using a deep network with parameters $\theta$.
    \item $p(y_\text{kp} | y) \propto \exp(-\lambda|y_{kp} - \mkp(y)|)$ where $\mu_\text{kp}(y)$ is the mean location of the keypoints estimated using a deep network with parameters $\theta_\text{kp}$ that takes part labels as input and predicts the locations of keypoints.
    \item $p(y_\text{mask} | y) = B(y_\text{mask},\mm(y))$ a Binomial distribution where $\mm(y)$ is obtained by summing over the parts probabilities at each pixel. This function has no learnable parameters.
\end{itemize}

\noindent
In the E Step we optimize $q(y)$ for each training example $x$ as: 
\begin{equation}
    \argmax_{q(y)} \sum_{y}  q(y)  \left[ \log p(y|x)\prod_{i=1}^n p(y_i|y) \right] + H(q).
\end{equation}
Given the form of the probability distributions this corresponds to maximizing $q(y)$ given $\mu(x), y_{kp}$ and $y_\text{mask}$ (ignoring the entropy term):

\begin{equation}
    \sum_{y}  q(y) \exp\big(-\alpha\big|y - \mu(x)\big|\big)  \exp\big(-\lambda\big|y_{kp} - \mu_\text{kp}(y)\big|\big) B\big(y_\text{mask},\mu_\text{mask}(y)\big).
\end{equation}

For hard EM,  it is possible to solve for the optimal $y$ using gradient ascent as each of these functions $\mu, \mu_\text{kp}(y)$ and $\mu_\text{mask}(y)$ are differentiable wrt $y$. 
Similarly, one can construct a sample estimate for $q(y)$ using gradient-based techniques such as SGLD~\cite{welling2011bayesian}.
However, both these choices require many gradient iterations and can get stuck in local minima as $y$ is very high-dimensional. Thus, instead of optimizing for each example $x$ individually we approximate the mode with another distribution $q_x(y) \approx q(y | x, y, y_{kp}, y_{mask}, \theta_q)$ parameterized using a deep network with parameters $\theta_q$ shared across all training examples. The network takes as input the image and coarse labels and predicts the part labels. In the E step we optimize $\theta_q$ using gradient descent over \emph{all} examples allowing us to amortize the inference cost across examples.

In the M step, for each unlabeled image $x$ we sample labels $y$ using the variational distribution $q(y | x, y, y_{kp}, y_{mask}, \theta_q)$ and update the parameters $\theta$ and $\theta_{kp}$ of the model for predicting $p(y|x)$ and $p(y_{kp}|y)$ respectively. In practice we sample the mode of each input $x$ predicted by the feed-forward network. This is simple and has worked well for our experiments, though techniques for sampling from deep networks might lead to better estimates. The entire algorithm is outlined in Alg.~\ref{alg:em}. 
Here $\ell, \lkp$ and $\lm$ correspond to the loss functions for the part labels, keypoints, mask  obtained as the negative log-likelihood of the corresponding probability functions in Eqn.~\ref{eq:joint} and $\ell_q$ is the negative entropy.

\paragraph{\textbf{Remarks.}} (1) In the above derivation we assumed all the images have the same set of coarse labels. But the method can be generalized to handle images with different number of coarse labels as the log-likelihood (Eqn.~\ref{eq:joint}) decomposes over the labels. However, the model for estimating the variational distribution $q(y)$ should be adapted to condition on the provided labels for the image. One possibility is to train separate models, e.g., $q(y|x,y_{mask})$ and $q(y|x,y_{kp}, y_{mask})$, or treat the missing labels as latent variables and infer them during training. (2) The method can handle different styles of coarse supervision by simply adding $p(y_k|y)$ for the corresponding label style. For example, 
supervision from object bounding-boxes can be incorporated by treating the box as two keypoints corresponding to the top-left and bottom-right corners or as a mask. Similarly, box-level annotations for the parts can also be used as coarse supervision.

\floatname{algorithm}{Algorithm}
\renewcommand{\algorithmicrequire}{\textbf{Input:}}
\renewcommand{\algorithmicensure}{\textbf{Output:}}

\begin{algorithm}[t]
\caption{Stochastic Variational EM for Part Segmentation}
\label{alg:em}
\small
\begin{algorithmic}[1]
\Require ${\cal D}^p := \{(x^p, y^p, \ym^p, \ykp^p)\}$ \Comment{Dataset with part labels}
\Require ${\cal D} := \{(x, \ym, \ykp)\}$ \Comment{Dataset with coarse labels}
\Require {params = \{$\#epochs, b^p,b, \alpha, \lambda_1,\lambda_2, \delta_1, \delta_2\}$} 
\Function{TrainPartSegSGDVarEM}{${\cal D}^p,{\cal D}$, params}
\State{Initialize $f(y|x, \theta)$, $f(\ym|y)$, $f(\ykp|y, \tkp)$ and $f(y|x,\ym,\ykp,\tq)$}
\For{$epoch \gets 1$ to \#epochs}
 \State{$[x^p, y^p, \ym^p, \ykp^p]$ = $\text{next-batch}({\cal D}^p, b^p)$}
 \State{$[x, \ym, \ykp]$ = $\text{next-batch}({\cal D}, b)$}
 \State{\textbf{\#E Step}}
 \State{$\mu_q = f(y|x,\ym,\ykp,\tq)$} \Comment{Variational distribution}
 \State{$\mu = f(y|x, \theta)$}         \Comment{Part segmentation model}
 \State{$\mkp = f(\ykp|\mu_q, \tkp)$}.  \Comment{Keypoint model}
 \State{$\mm = f(\ym|\mu_q)$}           \Comment{Mask model}
 \State{$L = \alpha \ell(\mu_q, \mu) + \lambda_1\lkp(\ykp, \mkp) + \lambda_2\lm(\ym,\mm) + \ell_q(\mu_q)$}
 \State{$\text{gradient-update}(L, \tq)$}
 \State{\textbf{\#M Step}}
 \State{$\mu_q = f(y|x,\ym,\ykp,\tq)$}     \Comment{Sample labels}
 \State{$\mu^p = f(y|x^p, \theta)$}         \Comment{Part segmentation model}
 \State{$\mkp = f(\ykp|\mu_q, \tkp)$}  \Comment{Keypoint model}
 \State{$\text{gradient-update}(\delta_1\ell(y^p, \mu^p) + \delta_2\ell(\mu_q, \mu), \theta)$}
 \State{$\text{gradient-update}(\lkp(\ykp, \mkp), \tkp)$}
\EndFor
\EndFunction
\end{algorithmic}
\end{algorithm}

\section{Benchmarks for Evaluation}\label{sec:benchmark}

\begin{figure*}[t]
\centering
\includegraphics[width=\linewidth]{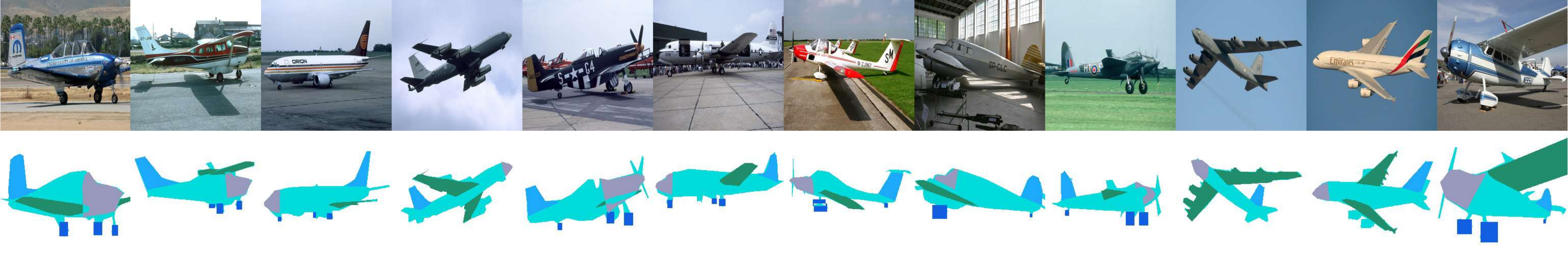}
\caption{\small{\textbf{Part labels on the OID Aircraft dataset.}}}
\label{fig:oid-examples}
\end{figure*}

In this section we describe the datasets used for our experiments. Fig.~\ref{fig:examples} shows the PASCUB dataset for bird part segmentation. The top row is examples we annotated from the CUB dataset and bottom row are examples from the PASCAL parts dataset for the birds category after removing low-resolution and truncated instances (Appendix A). Fig.~\ref{fig:oid-examples} shows examples from the OID Aircraft dataset. 
Below we describe the details and evaluation metrics of both.


\subsection{Bird part segmentation benchmark}
\label{sec:birddata}
Our goal is to segment each bird into 10 parts: `beak', `head', `left eye', `left leg', `left wing', `right eye', `right leg', `right wing', `tail' and `torso'. The bird category in the PASCAL parts dataset contains several labeled examples, but most instances are small and truncated as the dataset is primarily designed for object detection. On the other hand, the CUB dataset has higher resolution instances and includes keypoint and figure-ground masks but does not contain part labels. So, we combine the two and provide part labels for a few instances on the CUB dataset to create a benchmark for few-shot part segmentation.

\begin{flushleft}
\begin{minipage}{0.98\textwidth}
  \begin{minipage}[b]{0.26\textwidth}
    \centering
\resizebox{\textwidth}{!}{%
   \begin{tabular}{ccc}
\multicolumn{3}{c}{\textbf{Part Segmentation Data}} \\
\multicolumn{1}{c|}{Split} & PASCAL & CUB \\ \shline
\multicolumn{1}{c|}{\#Train} & 271 & 150 \\
\multicolumn{1}{c|}{\#Val} & 132 & 74 \\
\multicolumn{1}{c|}{\#Test} & 133 & 75 \\ 
\multicolumn{3}{l}{} \\
\multicolumn{3}{l}{\textbf{Coarsely Labelled Data}} \\\shline
\multicolumn{1}{c|}{\#Train} & \multicolumn{2}{c}{5994} \\
\multicolumn{1}{c|}{\#Val} & \multicolumn{2}{c}{2897} \\
\multicolumn{1}{c|}{\#Test} & \multicolumn{2}{c}{2897}
\end{tabular}}
      \captionof{table}{\small{\textbf{Data splits for PASCUB.}}}
      \label{tab:data}
    \end{minipage}
    \hfill
\begin{minipage}[b]{0.74\textwidth}
    \centering
    \includegraphics[width=\linewidth]{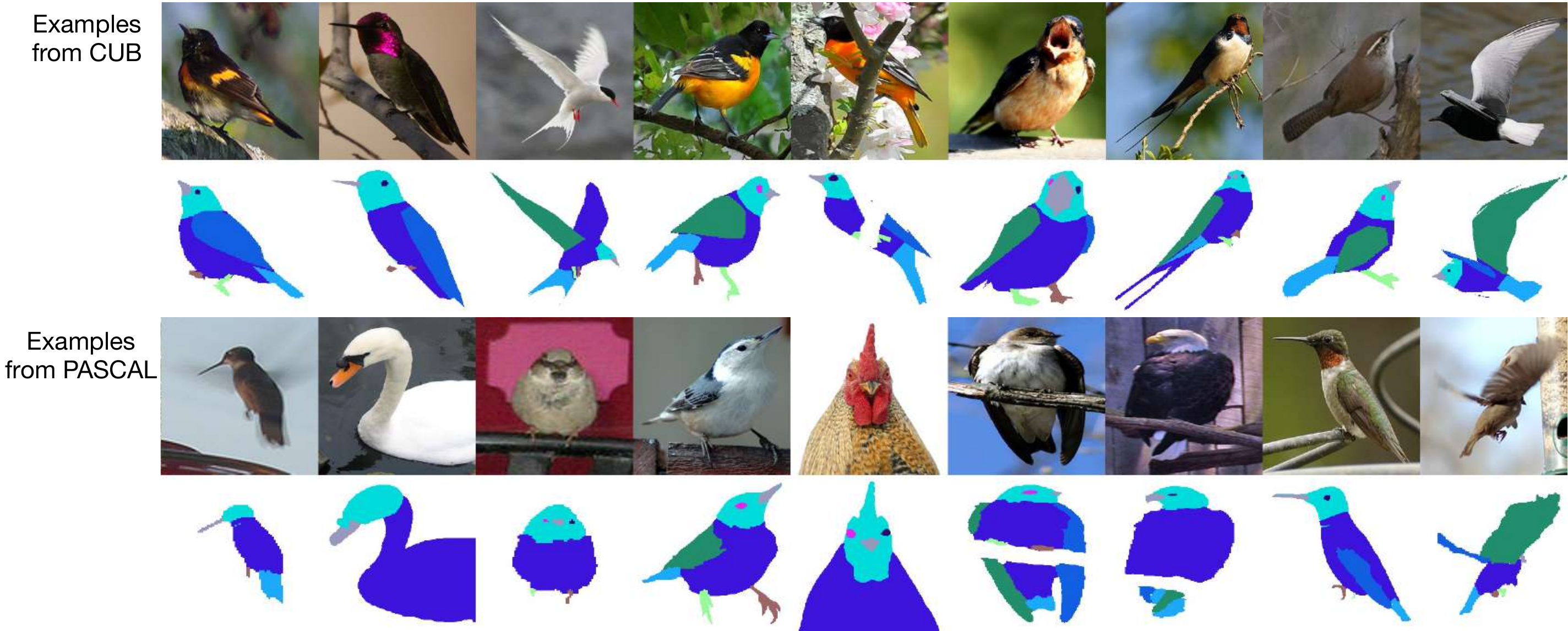}
    \captionof{figure}{\small{\textbf{Example images from the PASCUB dataset.}}}
    \label{fig:examples}
  \end{minipage}
  \end{minipage}
\end{flushleft}

\paragraph{\textbf{CUB}}
\label{sec:cub}
The Caltech-UCSD birds dataset~\cite{WahCUB_200_2011} has 11,788 images centered on individual birds across 200 species. We annotate 299 randomly chosen images with pixel-wise part labels (referred to as CUB Part) for the 10 classes mentioned above. We divide the 299 images we annotated into train, val and test in a 2:1:1 split (Tab.~\ref{tab:data}). The CUB dataset also includes keypoints and figure-ground masks for all images. We use the full data divided into the official splits as our coarsely labelled data for PASCUB experiments (Tab.~\ref{tab:data}).

\paragraph{\textbf{PASCAL}}
\label{sec:pascal}
The PASCAL VOC~\cite{Everingham10} dataset has 625 images that contain at least one bird. Chen \etal~\cite{chen2014detect} provide part segmentations where each bird has pixel-wise part labels for 13 classes -- we group classes such as `neck' and `head' to a single category `head' resulting in the 11 classes listed above. We also removed instances that are truncated and are of low resolution to make for a cleaner training and evaluation set ---  the pre-processing is detailed in Appendix A.  Now we are left with 536 centered bird images. Using the official split of PASCAL VOC results in a training set of 271 images. One image can contain more than one bird in PASCAL. Crops originating from an image from train split go in the train split of our dataset. Since the official split does not have val/test demarcation we randomly divide the rest of the images into validation and test sets equally (Tab.~\ref{tab:data}).
Fig.~\ref{fig:examples} shows the data after pre-processing.

The overall dataset contains 570 instances with part labels, and roughly 12k instances with keypoint and mask labels, divided into training, validation, and test sets. Annotating part segmentations requires roughly 5-10$\times$ more effort than masks based on our own experience, and this benchmark contains such labels for less than 5\% of the objects.

\subsection{Aircraft part segmentation benchmark}
The OID Aircraft dataset~\cite{mahendran14understanding} has 7543 images. Each image has an associated figure ground mask and part labels for four parts --- `nose', `wings', `wheels' and `vertical stabilizer'. The figure ground masks provided are quite accurate, but the part labels are noisy. Thus, we manually select 300 images for which the part labels are visually correct. In keeping with the splits of the datasets described above, we divide these 300 images into 150 for training, 75 each for validation and testing. We refer to this subset as OID Part. For the rest of the dataset, we use the official train/val/test split. Note that the part labels in this case do not collectively form the figure ground masks. Each pixel of the image also can have more than one part label marked. Thus, we handle the segmentation training in a different way for Aircrafts as described in the section below.



\section{Part Segmentation Algorithms}\label{sec:algorithms}
For all the baselines and for our approach, we use an encoder-decoder based fully convolutional network. We present details of all architectures in Appendix B. We use colour jittering and flipping augmentations for training all models. We resize images and corresponding labels to 256~$\times$~256. The hyperparameters for each approach were chosen on the validation set of each benchmark. 


\subsection{Baselines} \label{sec:baselines}
In this section, we describe training and design details for all baselines that we compare our method with.

\paragraph{\textbf{Fine-tuning}} We start with a network pre-trained on another task and fine-tune it for part segmentation. We replace the final fully connected layer to predict part labels and train the network using a cross entropy loss for PASCUB experiments. For Aircrafts, we treat segmentation as a pixel-wise multi-label classification task and use binary cross entropy (BCE) on each channel. We train this using Adam optimizer with a learning rate of 0.0001 for 200 epochs.


\paragraph{\textbf{Semi-supervised learning}} We use the method described in PseudoSup~\cite{chen2021semi} as a semi-supervised learning baseline. 
The method uses an ensemble of two networks obtained by fine-tuning starting from two different initializations. Note that for the `Random' case (see Tab.~\ref{tab:comparison}), both networks start with different random init before fine-tuning, while for `Keypoint'/`ImageNet' cases only the last layer/decoder has different random inits. After obtaining the two different fine-tuning checkpoints, PseudoSup training uses one ensemble to train the other and vice versa using pseudo-labels on all coarsely labelled images. Pseudo-labels refers to converting the predictions to one-hot labels by computing the argmax over all channels. We also add the fully-supervised loss from images with part labels. We use SGD optimizer with learning rate of 1E-4, momentum of 0.9 and weight decay of 1E-4 for both networks. We train for 90 epochs with cosine learning rate scheduling.


\paragraph{\textbf{Multi-task learning}} Here we train a single model to accomplish both the tasks of keypoint prediction and part segmentation. We use a common encoder based on a ResNet-34 and attach decoders for each task described below.
\begin{itemize}
    \item For PASCUB, the first decoder is for part segmentation labels where we use cross entropy loss over the prediction and ground truth labels. The second decoder predicts keypoints where we use pixel-wise $\ell_1$-loss over the predicted and ground-truth keypoints which are represented as Gaussians around each keypoint. The output of the first decoder also receives supervision from the figure-ground masks by summing over channel dimension for the foreground classes of the prediction and taking a cross-entropy loss. The sum of all these losses is backpropagated through the encoder during training. The weights for the figure-ground loss and part segmentation loss are set to be 1 and that of the keypoint loss is set to be 10. We use SGD optimizer with learning rate of 0.1, momentum of 0.9 and weight decay of 0.0001 for both networks. We train for 90 epochs with cosine learning rate scheduling.

    \item For Aircrafts, the first decoder performs part segmentation and is trained with binary cross-entropy loss for on each channel as parts are not mutually exclusive. The second decoder predicts the figure ground mask and is trained with cross-entropy loss. The weightages for losses from both decoders are set to be 1. Using an initial learning rate of 0.2, the rest of the training procedure remains the same as above.
\end{itemize}


\paragraph{\textbf{Handcrafted loss functions}} We base this method on PointSup~\cite{cheng2021pointly} --- a method to train segmentation models using point supervision. We evaluate this on PASCUB since it has keypoint annotations. The procedure is illustrated in Fig. 7 in the Appendix. We first assign keypoints to each part manually based on their co-occurrence, e.g. the `head', `crown' and `throat' keypoints are assigned the `head' part. We then dilate these locations using a 5$\times$5 pixel window --- we choose 5$\times$5 so as to not exceed the area of the smallest part, the eyes. We then train the network with a pixel-wise cross entropy loss computed on all these annotated points and the corresponding figure-ground mask. This is mask-loss is computed across all pixels by summing over the foreground channels and using a cross entropy loss. For the loss over part labels points we set the weightage as 0.5, for loss from figure-ground mask as 1 and for that from part label we set weightage to 2. We use SGD optimizer with learning rate of 0.001, momentum of 0.9 and weight decay of 0.0001. We train for 90 epochs with cosine lr scheduling. 


\subsection{Details for our approach}
\label{sec:init}

In this section we specify how we initialize each model of the EM algorithm before training and describe the training details of the EM method.

\paragraph{\textbf{Part segmentation model:}} $f(y|x)$. We initialize this using a checkpoint obtained by fine-tuning, i.e., a model trained using the provided part labels.

    

\paragraph{\textbf{Posterior inference model:}} $f(y|x,\ym, \ykp)$. We use a split encoders for this model (Fig.5 in supp.). The first is a ResNet34 pretrained on ImageNet~\cite{ILSVRC15} to extract features from the image and second a shallow ResNet-based encoder to process the masks and keypoint heatmaps concatenated in channel dimension. We concatenate the features of the encoders and use a common decoder to create $y$. For PASCUB, we train using images from CUB for which we have both labelled part segmentations and keypoint annotations. For Aircrafts, similarly we use those images which have both figure ground masks and clean part labels. We provide details on architecture in Appendix B. We use flipping and color jitter augmentations while training. We use a learning rate of 0.1 for the whole network except the image encoder branch for which we set learning rate to 0.01. We use cosine learning rate scheduling and train for 90 epochs. We use SGD optimizer with momentum of 0.9 and weight decay of 1e-4.

\paragraph{\textbf{Keypoint model:}} $f(\ykp|y)$. This refers to the model for predicting keypoints given part labels. On PASCUB using the checkpoint from the finetuned $p(y|x)$ model we generate part segmentations for all CUB images. We use $\ykp$ from ground truth and generated $y$ from $f(y|x)$ for an initial training. We then fine-tune the model on the images that have both ground truth $y$ and $\ykp$. For this stage we use color jitter and flipping augmentations. For the initial training we use a learning rate of 0.1 with cosine lr scheduling and train for 90 epochs. For fine-tuning, we use a learning rate of 0.001 and train for 10 epochs. We use SGD optimizer with momentum of 0.9 and weight decay of 1e-4 for both. This model achieves a PCK@10\% of \textbf{92.85} on the CUB test set which is higher than that obtained by training using image inputs (92.65) for the same architecture.

\paragraph{\textbf{Mask model:}} ${f(\ym|y)}$. This is the model for predicting figure-ground masks from part labels. For PASCUB, we can predict the mask directly by marginalizing (summing) over the all the part labels. For Aircrafts, we need to use a model to predict $f(\ym|y)$ since the part labels do not cover the whole mask and are not mutually exclusive. We use a model similar to the $f(\ykp|y)$ for PASCUB and follow the same initialization procedure.

\paragraph{\textbf{EM Training}}
\label{sec:emtraining}
As described in Alg.~\ref{alg:em}, the EM training proceeds by updating the posterior model $p(y|x, \ykp,\ym)$ (E Step), followed by updating the part $p(y|x)$ and keypoint $p(\ykp|y)$ models (M Step) over batches of training data.
For PASCUB with keypoint and ImageNet initialization, we use learning rate for part model as 1e-5, that of posterior model as 1e-3 and coarse supervision model $p(y_k|y)$ as 1e-8. For random initialization, we set learning rates as 5e-4, 1e-5 and 1e-8 respectively. We use SGD optimizer with momentum of 0.9 and weight decay of 1e-4. We train the model for 40 epochs and choose the best checkpoint based on lowest cross entropy loss of $p(y|x)$ on validation set of PASCUB dataset.  We use batch size of 32 for the coarse labelled dataset and a batch size of 4 for the part labelled dataset. We detail rest of the hyperparameters in Appendix C. 
We follow a very similar procedure for the Aircrafts. The loss for the E step comes from the predicted labels and the figure-ground mask, while for the M step we train the part model $p(y|x)$ using posterior mode. 
We use learning rate for part model as 0.005, that of posterior model as 1e-6 and coarse supervision model $p(y_m|y)$ as 1e-8. For Aircrafts we perform experiments for ImageNet init and share details of all hyperparameters in Appendix C. Fig.~\ref{fig:em-training} shows the progress segmentation models using EM over epochs on an image.

\begin{figure*}
    \centering
    \includegraphics[width=0.9\linewidth]{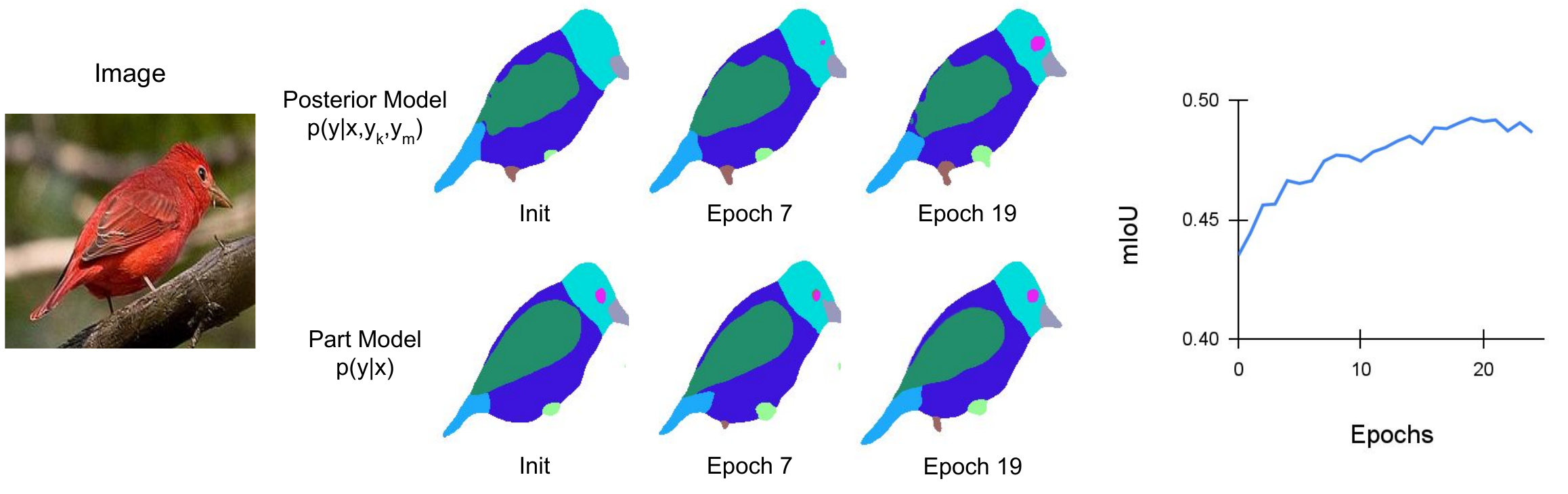}
    \caption{\small{\textbf{EM training.} The top row shows the output of the posterior model $p(y|x, \ym,\ykp)$ and the bottom shows the output of the part model $p(y|x)$ for the image at various epochs. Both models improve and influence each other -- the posterior model learns to recognise the eyes while the part model learns to segment the feet. The right shows that the validation mIoU increases over epochs. }
    }
    \label{fig:em-training}
\end{figure*}


\section{Results}\label{sec:results}
Below we summarize the key conclusions of our experiments.

\paragraph{\textbf{Our approach outperforms alternatives.}} Tab.~\ref{tab:comparison} compares our approach to baselines on the PASCUB dataset for various network initializations. Performance is reported as the mean intersection-over-union (mIoU) across parts.
Note that the benchmark has train/val sets for CUB and PASCAL. Tab.~\ref{tab:comparison} shows the results on CUB, while those for PASCAL are included in the supp. Our approach handily outperforms the fine-tuning baseline --- with the largest gains when the network is randomly initialized. We also outerform PointSup, a strong baseline based on handcrafted labels obtained from keypoints. Designing handcrafted labels might be challenging if keypoints are densely labeled, or if the annotation style varies. In comparison, our approach does not assume prior knowledge on the style of labels and learns them as part of training. Tab.~\ref{tab:oid} shows the same results for the OID Part dataset, where our approach outperforms fine-tuning and multi-tasking baseline. For this dataset we use ImageNet initialization. PointSup is not applicable as keypoints are not annotated on this dataset.


\paragraph{\textbf{Multi-tasking is rarely effective.}} The simple strategy of multi-tasking was effective only when the models are trained from scratch. Despite careful hyperparameter search, we found that the overall performance degrades when better initializations are used. A staged strategy, where the network is trained to predict keypoints on the whole CUB dataset and then fine-tuned to predict part labels was more effective (Tab.~\ref{tab:comparison} Keypoint init. + Fine-tuning outperforms Multi-task). 


\paragraph{\textbf{Semi-supervised learning provides minor benefits.}} The semi-supervised learning approach based on PseusoSup provides relatively small (0.5-1\% MIoU) improvement over the fine-tuning baseline.


\paragraph{\textbf{Our approach benefits from various coarse labels.}} Table~\ref{tab:emvariation} shows the results on the CUB test set using various forms of coarse supervision. A model trained using mask supervision only obtain 46.30\% mIoU, one with Keypoint only obtains 47.96\% mIoU, while using both Keypoints and masks obtains 49.25\% mIoU. All these models are better than the fine-tuning baseline (45.37\% mIoU) and the semi-supervised learning baseline (46.01\% mIoU). 


\paragraph{\textbf{Our approach is relatively efficient.}} The key benefit of our approach is that it is relatively efficient. First, we were able to utilize existing labels on PASCAL and CUB dataset to train the part segmentation model. Our model required labeling ~300 part labels on CUB, half of which were used for evaluation. 
Considering that it takes on the order of a minute or two to label parts for each instance, the ability to train part-segmentation models using existing coarse labels is a compelling alternative to labeling large datasets of parts. 
Second, the overall training for our approach (7.5 hr) is also a small factor increase over fine-tuning (1 hr), semi-supervised learning (6.2 hr), multitasking (4 hr) and PseudoSup (2.5 hr) on a single NVIDIA RTX8000 GPU.


\begin{table}[t]
\caption{\small{\textbf{Performance on Birds.} Comparison of the EM method with baselines described in \S~\ref{sec:baselines} on the testing and validation set of CUB parts. Our method (in green) outperforms baselines for all initializations. We present results on PASCAL val/test splits in Appendix D. The std-deviation over runs for Fine-tuning, MultiTask, PseudoSup and EM is $<\pm$1 mIoU. For PointSup the std-deviation is $\sim \pm$2 mIoU.}} 
\centering
{%
\setlength{\tabcolsep}{5pt}
\begin{tabular}{c|ccc|ccc}
\multirow{2}{*}{Method} & \multicolumn{3}{c|}{CUB Part Test} & \multicolumn{3}{c}{CUB Part Val} \\
 & \multicolumn{1}{c}{Random} & \multicolumn{1}{c}{Keypoint} & \multicolumn{1}{c|}{ImageNet} & \multicolumn{1}{c}{Random} & \multicolumn{1}{c}{Keypoint} & \multicolumn{1}{c}{ImageNet} \\ 
 \noalign{\smallskip}
 \shline
 \noalign{\smallskip}
Fine-tuning & 29.88 & 41.12 & 45.37 & 35.28 & 44.64 & 48.62 \\
MultiTask & 36.96 & 38.00 & 41.27 & 40.24 & 41.74 & 43.93 \\
PseudoSup~\cite{chen2021semi} & 30.77 & 41.62 & 46.01 & 36.32 & 45.03 & 48.67 \\
PointSup~\cite{cheng2021pointly} & 35.18 & 46.45 & 46.76 & 38.05 & 48.01 & 48.84 \\
Ours & {\color[HTML]{009901} 37.98} & {\color[HTML]{009901} 49.25} & {\color[HTML]{009901} 48.05} & {\color[HTML]{009901} 40.85} & {\color[HTML]{009901} 52.19} & {\color[HTML]{009901} 51.11}
\end{tabular}%
}
\label{tab:comparison}
\end{table}


\begin{table}
\RawFloats
\begin{minipage}{.45\linewidth}
\centering
\captionof{table}{\small{\textbf{Effect of coarse supervision.} The mIoU on the CUB test using various coarse labels.}
}
\setlength{\tabcolsep}{10pt}
\begin{tabular}{cc}
\noalign{\smallskip}
EM Supervision & mIOU \\ \shline
\noalign{\smallskip}
Keypoint + Mask & 49.25 \\
Mask only & 46.30 \\
Keypoint only & 47.96\\
\noalign{\smallskip}
\end{tabular}
\label{tab:emvariation}
\end{minipage}
\hfill
\begin{minipage}{.5\linewidth}
\centering
\captionof{table}{\small{\textbf{Performance on OID.} Our method (in green) outperforms baselines based on multi-tasking and fine-tuning.}
}
\setlength{\tabcolsep}{10pt}
\begin{tabular}{ccc}
\noalign{\smallskip}
Method & OID val & OID test \\ \shline
\noalign{\smallskip}
Fine-tune & 54.17 & 55.30 \\
MultiTask & 55.94 & 55.61 \\
Ours & {\color[HTML]{009901} 57.46} & {\color[HTML]{009901} 58.68}\\
\noalign{\smallskip}
\end{tabular}
\label{tab:oid}
\end{minipage}
\end{table}



\section{Conclusions}
We present a framework for learning part segmentation models using a few part labels by exploiting existing coarsely labelled datasets.
Our approach jointly learns the dependencies between labeling styles allowing supervision from diverse labels. This allowed us to train a bird part segmentation model by combining the part labels on PASCAL VOC with figure-ground mask and keypoint labels on CUB dataset. The model outperforms baselines based on fine-tuning, semi-supervised learning, multi-tasking, as well as learning with handcrafted labels and loss functions. We also presented results on the Aircraft dataset where we improve over the baselines.
Our framework can handle multiple types of annotations (e.g., boxes, keypoints, masks, etc.) providing a way to combine existing labels across datasets without requiring manual translation across styles. For example, we could combine annotations from the NABirds dataset~\cite{7298658} which contains keypoints and object bounding-box to improve results.

\paragraph{Acknowledgements.} The research is supported in part by NSF grants \#
1749833 and \#1908669. Our experiments were performed on the University of Massachusetts GPU cluster funded by the Mass. Technology Collaborative.
\clearpage
%
%
\bibliographystyle{splncs04}
\bibliography{egbib}

\clearpage
\appendix

\noindent{\Large \textbf{Appendix}}

\section{PASCAL Pre-Processing}
Birds in PASCAL Parts are segmented into 14 parts : `beak', `head', `left eye', `left foot', `left leg', `left wing', `neck', `right eye', `right foot', `right leg', `right wing', `tail' and `torso'. We perform the following pre-processing steps to create a dataset for bird part segmentation:
\begin{itemize}
    \item Crop individual objects from a full image based on annotated pixel-wise instance masks with a margin of 20 pixels on all sides
\item Drop the cropped image if part of any other instance is present in the crop
\item Check if the cropped instance is of bird class
\item Filter out crops that have no part segmentation annotation
\item Convert from 14 to 11 classes by dropping 'neck' class and merging 'leg'/'feet' classes due to inconsistencies in the annotated labels
\item Filter out crops which are too small i.e. if either height or width $\le$ 80 pixels
\item Filter out crops which contains mostly the head of the bird by calculating the ratio of area of body to head. If ratio is $\le$ 0.95, we drop the crop\footnote{Note that we use this area ratio based method as the 'truncation' and 'occlusion' labels of PASCAL Part dataset would leave out many useful images}.
\end{itemize}

After these pre-processing steps we have 536 centered bird images. Using the standard split of PASCAL VOC, we separate out a training set of 271 images. If the crop originates from a train set image it goes in the training set of our dataset. We randomly split the rest of the images into validation and test sets containing 132 and 133 images respectively.

\section{Architectures}

\begin{figure}[h]
    \centering
    \includegraphics[width=\linewidth]{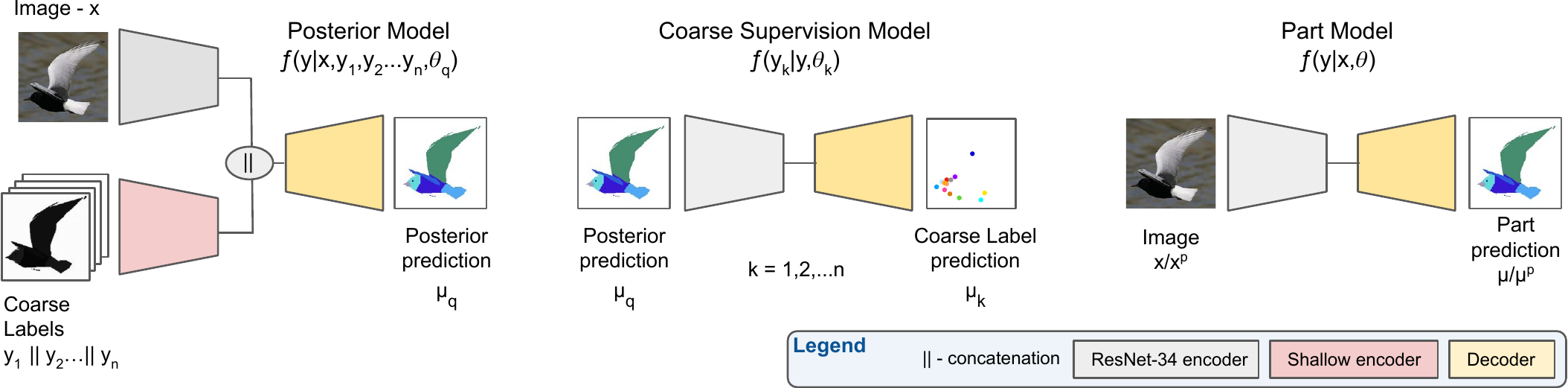}
    \caption{\small{\textbf{Architectures of networks used in the EM method.} This figure shows the input and output of each network. On the left is posterior model $f(y |x, \ym, \ykp)$, middle is  $f(y_k|y)$ and right is $f(y|x)$. Notations are consistent with Alg.~\ref{alg:em}. Note that $f(\ym|y)$ is deterministic, while  $f(\ykp|y)$ is parameterized as a deep network and trained as part of the EM algorithm (see \S~\ref{sec:init}).}
    } 
    
    \label{fig:init}
\end{figure}

\begin{figure*}[h]
\centering
\includegraphics[width=1\linewidth]{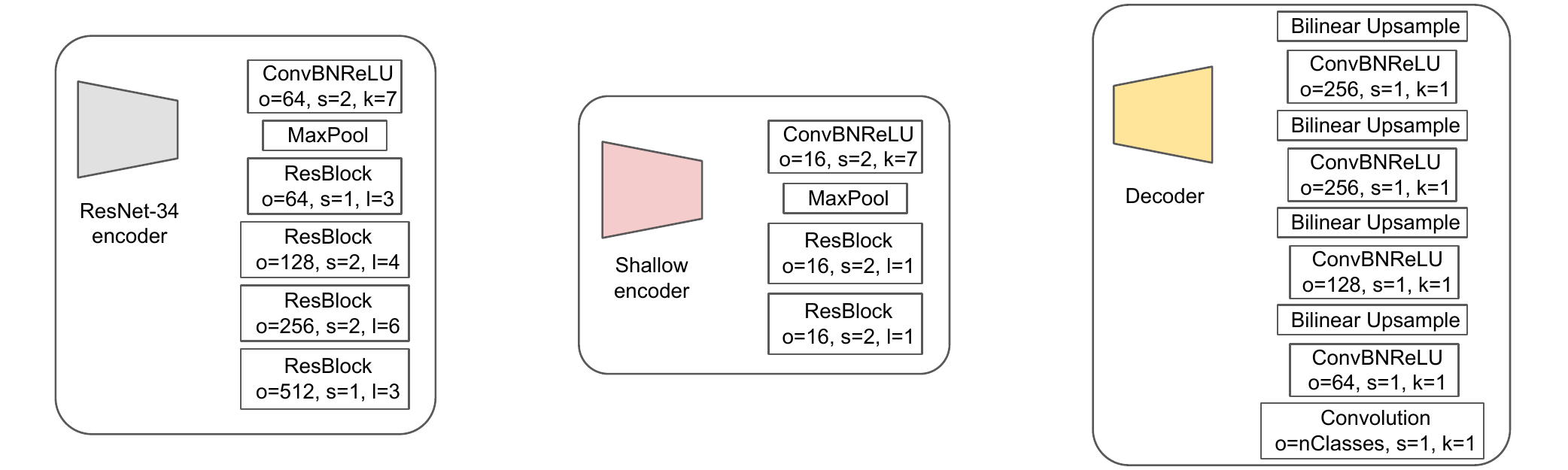}
\caption{\small{\textbf{Architectures in detail.} Here we describe the model architectures used in the EM algorithm shown briefly in Fig.~\ref{fig:init}. ResBlock refers to the Basic Block used by ResNet architectures~\cite{he2016deep}.} }
\label{fig:detail}
\end{figure*}

Note that the architectures used for the baselines are derivative of these architectures. The PointSup is made of the `ResNet-34 encoder' and `Decoder'. The MultiTask is made of one `ResNet-34 encoder' and two `Decoder's. The Fine-tuning and PsuedoSup baselines use `ResNet-34 encoder' followed by `Decoder'.

\section{Hyperparameters}

Here we list the rest of the hyperparameters for EM training. The loss term for E Step is $\alpha \ell(\mu_q, \mu) + \lambda_1\lkp(\ykp, \mkp) + \lambda_2\lm(\ym,\mm) + \gamma\ell_q(\mu_q)$. The loss for the part segmentation model is $\delta_1\ell(y^p, \mu^p) + \delta_2\ell(\mu_q, \mu)$.
For ImageNet init, we set $\alpha = 0.05, \lambda_1 = 50, \lambda_2 = 1, \gamma = 0.01, \delta_1 = 0.05, \delta_2 = 50$. For keypoint init, $\alpha = 0.05, \lambda_1 = 100, \lambda_2 = 1, \gamma = 0.01, \delta_1 = 0.001, \delta_2 = 50$. For random init, $\alpha = 0.01, \lambda_1 = 50, \lambda_2 = 1, \gamma = 0.01, \delta_1 = 0.1, \delta_2 = 100$. For training on the Aircrafts dataset, we set $\alpha = 0.05, \lambda_2 = 5, \gamma = 0.01, \delta_1 = 0.01, \delta_2 = 50$. For tuning the hyperparameters the relative values of $\alpha$ and $\lambda_1$ ($\lambda_2$ for Aircrafts) is important. Similarly for $\delta_1$ and $\delta_2$. For $\alpha$ we sweep from [0.001,0.005,0.01,0.05,0.1,0.5], for $\lambda_1$:  [10,50,100,500], for $\delta_1$:  [0.001,0.005,0.01,0.05,0.1,0.5], for $\delta_2$: [10,50,100,500]. We keep $\gamma$ as 0.01 for all cases.

\section{Results on PASCAL Val/Test Splits}

In Table~\ref{tab:pascal} we present the quantitative performance of models trained with ImageNet initialization on validation and testing sets of PASCAL Birds. A very similar trend as CUB dataset follows for PASCAL too.
\begin{table}[h]
\caption{\textbf{Results on PASCAL Val/Test.}}
\centering
{%
\setlength{\tabcolsep}{5pt}
\begin{tabular}{c|cc}

Method     & PASCAL Test & PASCAL Val \\
 \noalign{\smallskip}
 \shline
 \noalign{\smallskip}
Finetuning & 34.25     & 32.02     \\
Multitask  & 30.34     & 29.50     \\
PseudoSup  & 34.98     & 33.28     \\
PointSup   & 35.72     & 35.78     \\
Ours         & {\color[HTML]{009901} 36.31}     & {\color[HTML]{009901} 37.52}   
\end{tabular}%
}\label{tab:pascal}
\end{table}

\begin{figure*}[t]
\centering
\includegraphics[width=\linewidth]{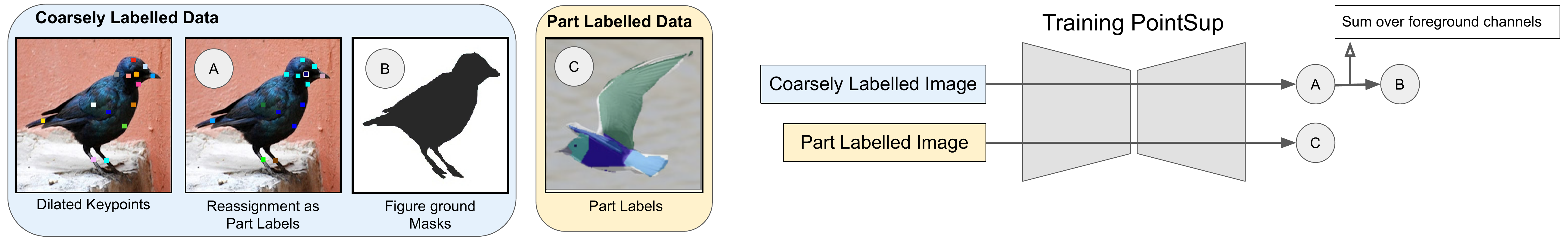}
\caption{\small{\textbf{Training PointSup} using figure ground masks and handcrafted part labels for coarsely labelled data, and using part labels for part segmented data}}
\label{fig:pointsup}
\end{figure*}

\newpage
\section{Qualitative Results}
\begin{figure}[!h]
\centering
\includegraphics[width=1\linewidth]{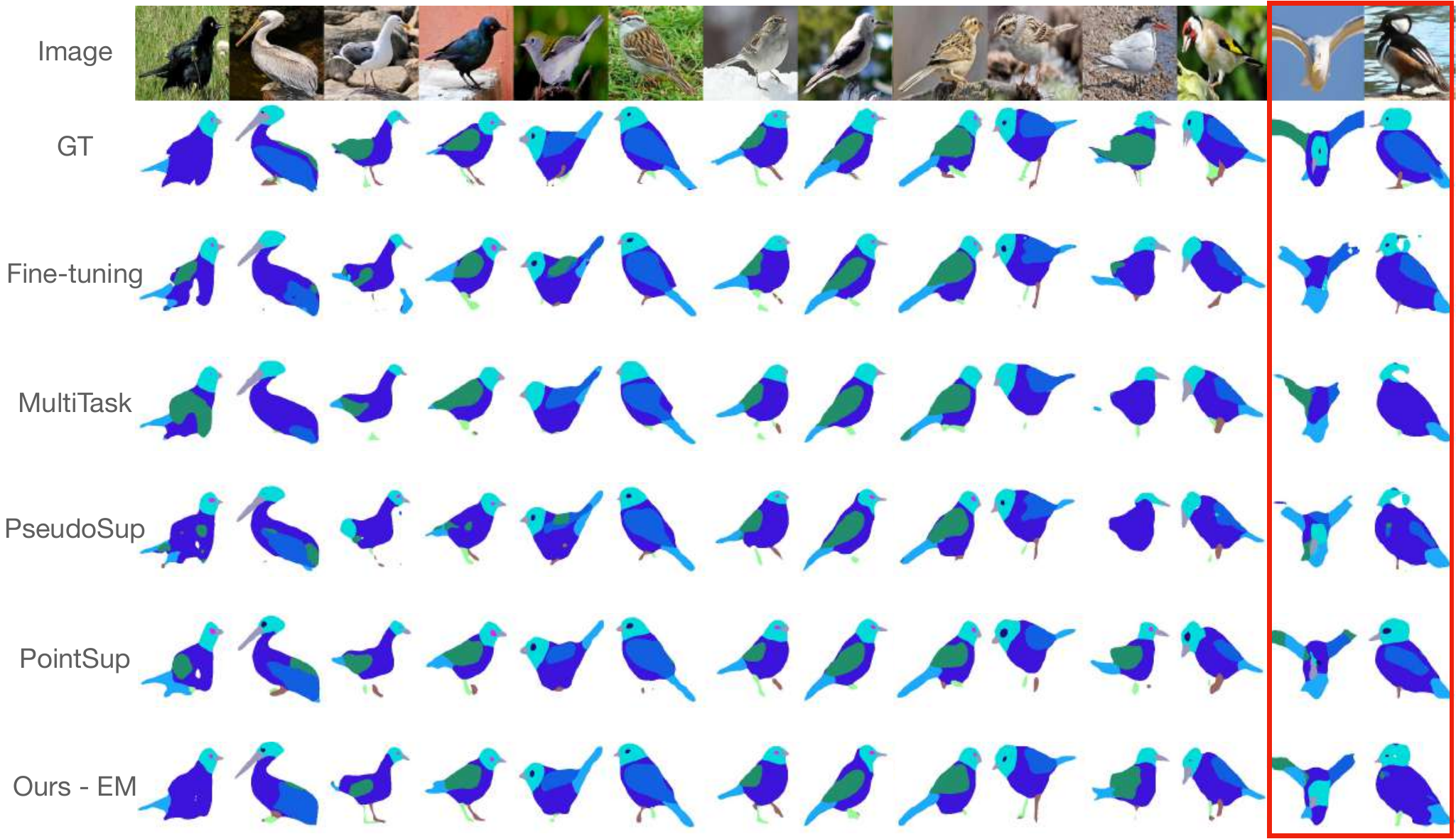}
\caption{\small{\textbf{Qualitative Results on PASCUB dataset.} The examples bordered in red show two of the failure cases.} }
\label{fig:birdviz}
\end{figure}

\begin{figure}[t]
\centering
\includegraphics[width=1\linewidth]{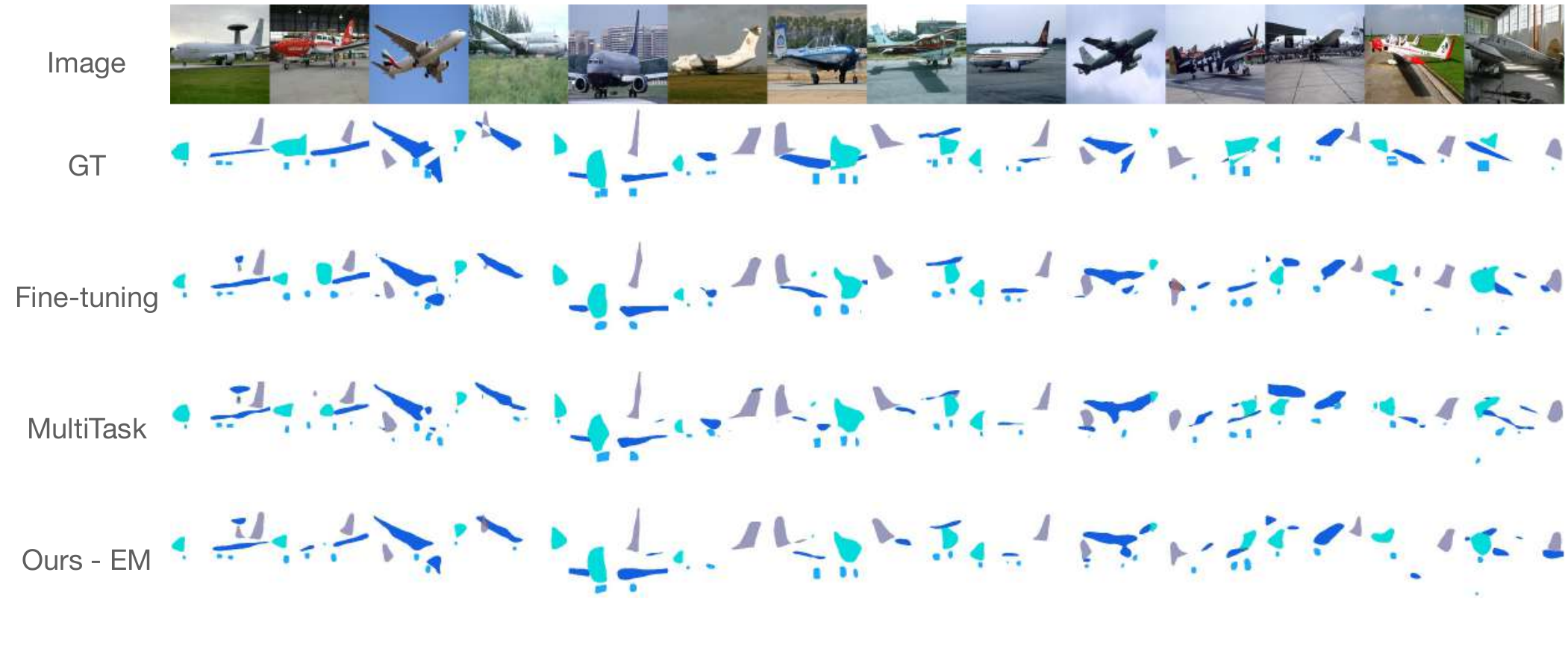}
\caption{\small{\textbf{Qualitative Results on OID Aircraft dataset.}} }
\label{fig:aircraftviz}
\end{figure}

\end{document}